\documentclass{article}

\usepackage{microtype}
\usepackage{graphicx}
\usepackage{subfigure}
\usepackage{booktabs} 
\usepackage{tikz}
\usepackage{caption}

\usepackage{xr-hyper}
\usepackage{hyperref}

\makeatletter
\newcommand*{\addFileDependency}[1]{
  \typeout{(#1)}
  \@addtofilelist{#1}
  \IfFileExists{#1}{}{\typeout{No file #1.}}
}
\makeatother
 
\newcommand*{\myexternaldocument}[1]{%
    \externaldocument{#1}%
    \addFileDependency{#1.tex}%
    \addFileDependency{#1.aux}%
}

\myexternaldocument{supmat}

\usepackage[accepted]{icml2019}

\icmltitlerunning{Using Deep Networks and Transfer Learning to Address Disinformation}

\begin{document}

\twocolumn[
\icmltitle{Using Deep Networks and Transfer Learning to Address Disinformation}



\icmlsetsymbol{equal}{*}

\begin{icmlauthorlist}
\icmlauthor{Numa Dhamani}{nk}
\icmlauthor{Paul Azunre}{alg}
\icmlauthor{Jeffrey L. Gleason}{nk}
\icmlauthor{Craig Corcoran}{nk}
\icmlauthor{Garrett Honke}{bng}
\icmlauthor{Steve Kramer}{nk}
\icmlauthor{Jonathon Morgan}{nk}
\end{icmlauthorlist}

\icmlaffiliation{nk}{New Knowledge, Austin, Texas, USA}
\icmlaffiliation{bng}{Watson School of Engineering and Applied Science and
the Department of Psychology: Cognitive and Brain Sciences, Binghamton University (SUNY), Binghamton, New York, USA}
\icmlaffiliation{alg}{Algorine, Inc., Austin, Texas, USA}

\icmlcorrespondingauthor{Numa Dhamani}{numa@newknowledge.io}

\icmlkeywords{NLP, CNN, LSTM, text classification}

\vskip 0.3in
]



\printAffiliationsAndNotice{}  

\begin{abstract}
We apply an ensemble pipeline composed of a character-level convolutional neural network (CNN) and a long short-term memory (LSTM) as a general tool for addressing a range of disinformation problems. We also demonstrate the ability to use this architecture to transfer knowledge from labeled data in one domain to related (supervised and unsupervised) tasks. Character-level neural networks and transfer learning are particularly valuable tools in the disinformation space because of the messy nature of social media, lack of labeled data, and the multi-channel tactics of influence campaigns. We demonstrate their effectiveness in several tasks relevant for detecting disinformation: spam emails, review bombing, political sentiment, and conversation clustering.
\end{abstract}

\section{Introduction}

Electronic communication is more embedded and essential to human life than ever before. This communication increasingly relies on the distributed, self-publishing model of social media platforms. The increasing ease of distributed communication does not come without drawbacks: disinformation---deceptive information spread deliberately to change behavior, influence public opinion, or obscure the truth---has infiltrated the online information ecosystem, with damaging consequences \citep{3, 4}. 

Malicious electronic communication takes many forms. It spans from low-level social engineering attacks (i.e., phishing) to more sophisticated, distributed efforts to disseminate state propaganda \citep{inkster2016, Okoro2013, Woolley2016}. We propose that the language characteristics of known and identified sources of malicious electronic communication can be used as a signal for the detection and mitigation of these efforts across the diverse (and fractured) electronic communication ecosystem. 

The motivation of this work is to demonstrate how semantic classification of natural language can be used as a tool for the detection of inflammatory, inauthentic, or otherwise nefarious communication. Character-level convolutional neural networks (CNNs) are particularly well-suited for this task---as opposed to a word-level model---because they allow for non-vernacular discourse, misspelling, and other social media features (e.g., emoticons) to be learned without the constraint of fixed vocabularies \cite{1}. We implement an adaptation of a neural network architecture recently demonstrated to be effective for text classification \cite{1, JozefowiczLimits16}. The method is purely content-based and does not require any additional metadata beyond the text. To show the effectiveness of this method in relation to malicious communication and disinformation, we present a series of experimental results on semantic classification for spam emails, review bombing, political sentiment, and conversation clustering.

\section{Related Work}

The way people consume and produce information online looks radically different today than it did in the recent past \citep{5}. This change has introduced a societal-level vulnerability---where foreign entities have been accused of interfering in the operations of sovereign democracies \citep{inkster2016, 17, Woolley2016, 19}. Even internally in various states, political differences have encouraged inauthentic, organized disinformation campaigns to attack brands \citep{20, 21}. This situation has led to an increased interest in studying the dissemination of inauthentic and/or false information to find solutions that protect the integrity of online discourse and the democratic institutions that depend on it. We believe that a computational technique that can use the properties of electronic discourse for the purpose of identifying key characteristics (e.g., source, intent, and effect) can be effective for addressing this complex problem.

Few frameworks have been proposed to detect and/or monitor malicious communication. One existing approach is \textsc{hoaxy}, a platform designed to collect, detect, and visualize the online spread of misinformation and fact-checking \citep{6}. \textsc{truthy} (another approach) uses network analysis techniques to track political memes for the purpose of detecting misinformation in U.S. political elections \citep{7}. It includes an automated system that assesses the credibility of posts on social media using content-based features and user metadata \citep{8}.

Motivated by computational journalism, \textsc{factwatcher} helps journalists identify newsworthy facts supported by data that can serve as leading news stories \citep{12}. Another attempt to fact-check simple statements employs the use of a public knowledge graph extracted from Wikipedia, framing the problem in terms of network analysis \citep{11}. There have also been efforts to create automated fact-checking systems for multimedia, where the goal is to distinguish between fake and real images using classification models based on user features and post content \citep{9, 10}. Platforms that focus on the veracity of claims more broadly---e.g., \textsc{politifact.com} and \textsc{factcheck.org}---rate the accuracy of claims made in political debates, speeches, and ads. More recently, computational resources and models to detect fake news have been developed which rely on linguistic features \citep{27}.

Current research in this domain includes tracking and analyzing the diffusion of rumors. \textsc{rumorlens} is a suite of interactive tools to help identify rumors on social media and analyze their impact \citep{13}. Similarly, \textsc{twittertrails} allows users to automatically answer questions about rumors on Twitter (e.g., origin, propagators, main actors, etc.) and compute level of visibility \citep{14}. Another rumor-checking website is \textsc{snopes.com} which focuses on urban legends and hoaxes.

\section{Current Approach} \label{Approach}

The work surveyed above is notable for its goal of addressing a difficult problem with the use of all possible features. In the same sense, however, a limitation of this work is its reliance on complete and detailed information from the platform (i.e., most often the \textsc{twitter} platform, which makes available metadata and network graph characteristics). There is a current need, therefore, for an approach that can successfully classify content of interest across platforms, with only the text content as input and in the absence of any exogenous features---features that cannot be relied on consistently in cross-platform analysis. Our proposed method is a multi-label, multi-class semantic classifier able to successfully handle natural language across platforms. The capability to handle naturalistic social language---e.g., emoticons, slang, misspellings---at the character-level gives this framework potential to be a powerful tool in detecting malicious communication from peer-to-peer social engineering to distributed disinformation across the electronic communication ecosystem.

\subsection{Neural Network Architecture Overview}

Depictions of the architectural inspiration for the current approach\footnote{Source code: https://github.com/NewKnowledge/simon} are presented in Figure \ref{model2} (see \citeauthor{1}, \citeyear{1}; \citeauthor{JozefowiczLimits16}, \citeyear{JozefowiczLimits16} for in-depth architecture information). It consists of two coupled deep neural networks---a network that encodes each individual sentence and a network that classifies the entire document.

\begin{figure}[hbtp!]
    \centering
    \includegraphics[width=0.5\textwidth]{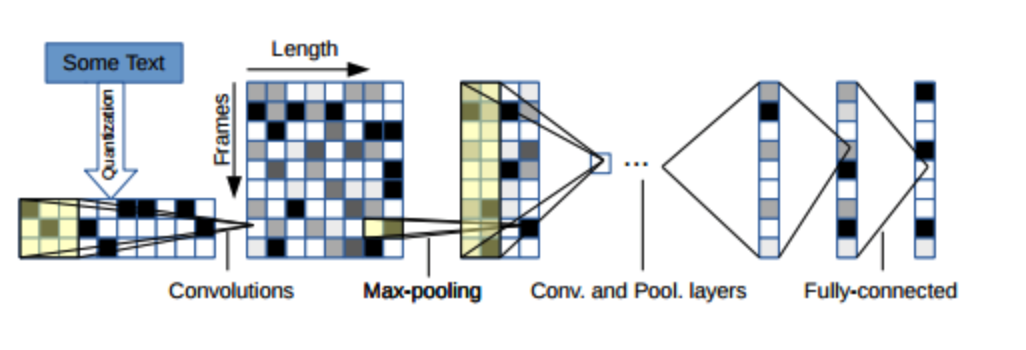}
    \caption{Diagram of character-level CNN model architecture. Figure reproduced from \citeauthor{1}, \citeyear{1}.}
    \label{model1}
\end{figure}

\begin{figure}[hbtp!]
    \centering
    \includegraphics[width=0.5\textwidth]{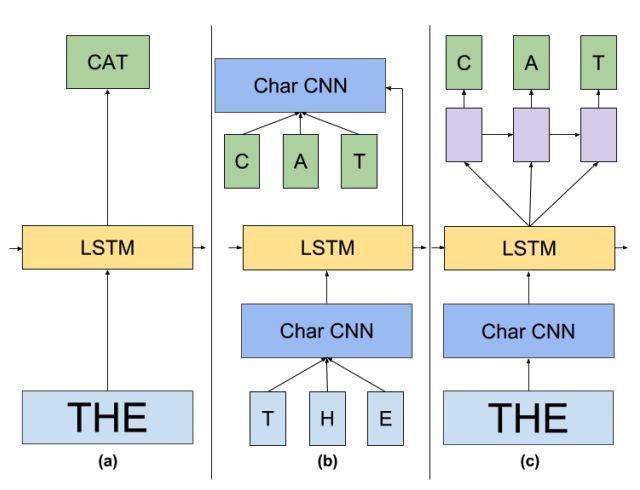}
    \caption{Word- and character-level models for text generation. This work uses the encoder from (b). Figure reproduced from \citeauthor{JozefowiczLimits16}, \citeyear{JozefowiczLimits16}.}
    \label{model2}
\end{figure}

The first network consists of 13 convolutional, max-pooling, dropout and bidirectional LSTM layers. Characters of an input sentence are one-hot encoded from a dictionary of 71 characters and bounded by a maximum length (tunable, depending on the application). It produces the final sentence encoding (a 512-dimensional feature vector).

The second network, which classifies the document as a whole, consists of 7 convolutional, max-pooling, dropout, and bidirectional LSTM layers. The input is again bounded by a maximum number of sentences (a tunable parameter). The final layer outputs the probabilities of classifying the document as each different type of class.

\section{Empirical Assessment} \label{Empirical Assessment}

Three examples are provided below to showcase how the CNN--LSTM classification approach can be applied to domains of interest. The final section describes how the process can be generalized to conduct unsupervised conversation clustering. The procedure is similar across applications: (1) initial weights are learned with training on the language(s) of the target domain and---if necessary---(2) transfer learning with labeled data is performed to learn the specific classes of interest (e.g., spam vs. ham, on-topic vs. off-topic review content, etc.).

\subsection{Spam Email}

Peer-to-peer social engineering attacks constitute a major threat vector at individual, business, and geopolitical levels. Perhaps the most benign example of this threat is commercial spam. Thus, a good starting point for assessment is testing the applicability of the CNN--LSTM architecture to this problem. In addition to being able to classify an email as spam or authentic, this use case can be extended to build a multi-label, multi-class classifier that differentiates between a broader set of attacks, e.g., propaganda and social engineering emails vs. phishing. We employ the transfer-learning paradigm to adapt the initial classes to a more sophisticated problem---first learning to determine if the originator of a piece of electronic communication is a friend or foe, and then learning to identify the specific class of adversarial communication best attributed to the document (see Figure \ref{fig1} in the \nameref{supmat}).

The classifier is initially trained on the popular Enron email dataset \footnote{\url{www.kaggle.com/wcukierski/enron-email-dataset}}, the 419 spam fraud corpus \footnote{\url{www.kaggle.com/rtatman/fraudulent-email-corpus}}, and an email abuse dataset acquired from NASA Jet Propulsion Laboratory (JPL). A dataset balanced between two classes (\emph{friend} and \emph{foe}) was generated with 7000 samples of each class. The model converged with a test binary accuracy of 96.14\%.

Then transfer learning is applied to the original binary classifier to complete the training of a full multi-class classifier, augmenting the number of handled classes from 2 to 8 with reduced labeled data and computing power requirements. The 8 classes are: friend, 419 scam, malware, credential phishing, phishing training, propaganda, social engineering, and spam. As shown in Figure \ref{fig1} in the \nameref{supmat}, the accuracy convergences quickly to a test accuracy of 93.25\%. Corresponding precision was calculated as 0.96, recall as 0.49, and F1 score as 0.65. This performance, using email text {\bf only}, provides further evidence of the effectiveness of transfer learning in natural language processing (NLP), which still remains relatively under-explored in this space \citep{15}.

\subsection{Review Bombing}

An important use-case for combating disinformation is the ability to detect inauthentic behaviors and campaigns used to threaten the integrity and good reputation of a product, company, or brand. For example, the perceived quality of products on marketplace-based websites can be manipulated through fake reviews \citep{22}. Likewise, movie reviews can be manipulated through coordinated campaigns that have the goal to negatively impact box office numbers and revenue---so-called \emph{review bombing}. One approach to combating this manipulation is to identify off-topic reviews, reviews that are not relevant to the product but take on peripheral issues (e.g., political stances of organizations or athletic endorsers).

To test the usefulness of CNN--LSTM based semantic classification in this domain, our method was trained on \emph{on-topic} vs. \emph{off-topic} movie reviews collected from a popular, crowd-sourced review website. The initial convergence results achieve a binary test accuracy of 99.5\%. These results suggest that this technique is a promising approach for detecting inauthentic behaviors for manipulation of products and brands.

\subsection{Political Sentiment}

Due to the central role social media platforms fill as critical communication infrastructure, malicious actors now have the ability to construct and disseminate false narratives with broad implications for society---often by piggybacking on polarizing content or current events. This behavior has significant and harmful effects \citep{26}, muddying the ability to use social media as a means to understand broad trends in preferences and behavior. To be able to combat fraudulent or misleading narratives, it is critical to be able to characterize the content as intentionally harmful (negative/anti), intentionally favorable (positive/pro), or neutral. 

Semantic classification was applied to posts discussing Howard Schultz' potential presidential election campaign using the text of the post {\bf only}. A three-class classifier---\emph{Pro}, \emph{Anti}, or \emph{Neutral}---was used. From a corpus containing approximately 6,000 supportive posts, 20,000 negative or harmful posts, and 40,000 neutral posts, we sampled 4,000 posts from each category. 

Figure \ref{fig2} in the \nameref{supmat} presents the training results on the political sentiment task. The model attained a binary test accuracy of 88.10\%. Notably, the accuracy of the classification of negative and/or harmful posts is perfectly (100\%) accurate. Corresponding precision, at a threshold probability of 0.5, was calculated as 0.68, recall as 0.98, and F1 score as 0.81. The ROC curve is also presented in Figure \ref{fig2}. We note that no hyper-parameter tuning was performed, which suggests that these metrics are amenable to further improvement.

\subsection{Conversation Clustering}

The current approach can also learn features that are useful for exploratory data analysis. The ability to examine the structure and trends of online discourse provides critical context for detecting and understanding the large-scale disinformation campaigns. We use the latent space generated by a pre-trained model to cluster conversations and visualize the results.

\begin{figure}[hbtp!]
    \centering
        \includegraphics[width=0.4\textwidth]{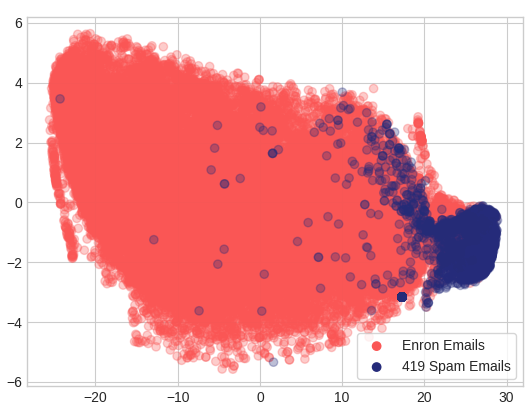}
        \caption{t-SNE embedding on Enron and 419 spam email datasets.}
    \label{fig3}
\end{figure}

An example is shown in Figure \ref{fig3}, which depicts a \emph{t}-Distributed Stochastic Neighbor Embedding (\emph{t}-SNE) clustering of the 128-dimensional features that the model learned from training on the Enron email dataset and the 419 spam fraud corpus. The features effectively separate the two classes and illustrates the increased variability in the Enron class as opposed to the spam class.

\section{General Discussion}

\subsection{Limitations and Alternatives}

One criticism of this work is that a simpler approach to text classification might be capable of achieving similar levels of performance. Indeed, methods such as bag-of-words and bag-of-\textit{n}grams models have comparable performance \cite{1} on many supervised tasks. However, we assert that the latent representation learned by the proposed approach better captures nuance in discourse than the bag-of-\textit{n}grams representation and, consequently, is better suited for transfer learning and exploratory tasks. The choice of a character-level CNN over the common approach of using word embedding spaces allows for misspelling, slang, unconventional characters (e.g., emoticons), and anomalous or unique vocabulary usage. These advantages are uniquely suited for the purpose of analyzing and tracking discourse patterns that are not represented in mainstream discourse.

\subsection{Ongoing Work}
The applications above begin to address small parts of the disinformation problem. We think of these as pieces of a broader approach necessary to effectively combat disinformation. Towards that goal, we have an ongoing effort that examines the dynamics of language usage across communities and social platforms. The kind of document embedding used above for transfer learning and clustering can also be used to measure the similarity of language use across communities, track changes in language use over time, and identify potential sources of those changes (e.g. adopting slang present in another platform or community, indicating a diffusion of ideas between the two groups).

\subsection{Conclusion}
We present a character-level CNN--LSTM model as an effective general-purpose tool for exploration and inference tasks relevant to disinformation on social media. As demonstrated in Section \ref{Empirical Assessment}, the model is able to learn a generic vector-space representation from a rich labeled dataset, then be utilized in label-scarce settings (common in the disinformation space) via transfer learning. The representation is also useful in an exploratory setting, which is important for understanding and adapting to the rapidly changing social media ecosystem. The flexibility of this framework to create a compact representation of natural language across a variety of communication channels by analyzing character-to-character contingencies makes it a powerful tool with broad potential to affect the quality and integrity of online discourse.

\subsubsection*{Acknowledgments}

Work was supported by the Defense Advanced Research Projects Agency (DARPA) under D3M (FA8750-17-C-0094) and ASED (FA8650-18-C-7889). Views, opinions, and findings contained in this report are those of the authors and should not be construed as an official Department of Defense position, policy, or decision.

\bibliography{icml2019}
\bibliographystyle{icml2019}

\onecolumn

\section*{Supplementary Material} \label{supmat}

\begin{figure}[hbtp!]
    \makebox[\textwidth][c]{\resizebox{1.00\textwidth}{!}{\input{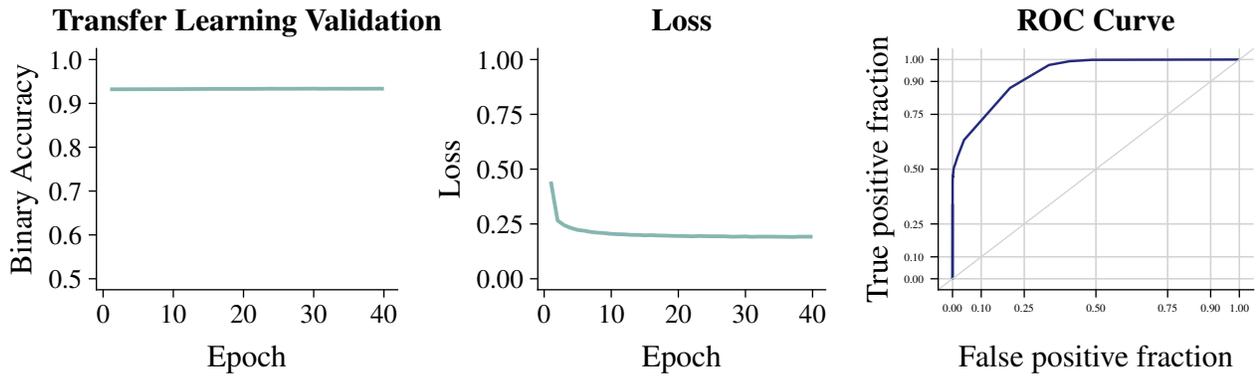}}}
    \caption{Convergence and performance results for the email spam classification task.}
  \label{fig1}
\end{figure}

\begin{figure}[hbtp!]
    \makebox[\textwidth][c]{\resizebox{1.00\textwidth}{!}{\input{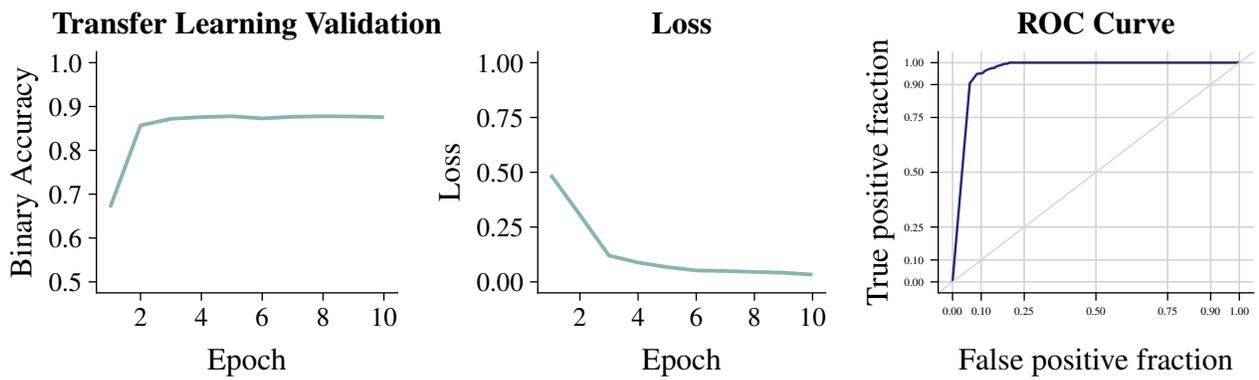}}}
    \caption{Convergence and performance results for the narrative categorization task.}
  \label{fig2}
\end{figure}

\end{document}